\documentclass[letterpaper]{article}

\usepackage{PRIMEarxiv}
\usepackage[english]{babel}
\usepackage{amsmath}
\usepackage{graphicx}
\usepackage[colorlinks=true,allcolors=blue]{hyperref}
\usepackage{booktabs}

\usepackage[round,authoryear]{natbib}
\usepackage{xcolor}
\usepackage{float}
\usepackage{etoolbox}
\usepackage{authblk}

\usepackage{eso-pic}

\newcommand{\model}{BDH-CQ}
\newcommand{\arcone}{ARC-AGI-1}
\usepackage[section]{placeins} %

\makeatletter
\newcommand{\mathleft}{\@fleqntrue\@mathmargin\parindent}
\newcommand{\mathcenter}{\@fleqnfalse}
\patchcmd{\NAT@citex}
  {\@citea\NAT@hyper@{%
     \NAT@nmfmt{\NAT@nm}%
     \hyper@natlinkbreak{\NAT@aysep\NAT@spacechar}{\@citeb\@extra@b@citeb}%
     \NAT@date}}
  {\@citea\NAT@nmfmt{\NAT@nm}%
   \NAT@aysep\NAT@spacechar\NAT@hyper@{\NAT@date}}{}{}
\patchcmd{\NAT@citex}
  {\@citea\NAT@hyper@{%
     \NAT@nmfmt{\NAT@nm}%
     \hyper@natlinkbreak{\NAT@spacechar\NAT@@open\if*#1*\else#1\NAT@spacechar\fi}%
       {\@citeb\@extra@b@citeb}%
     \NAT@date}}
  {\@citea\NAT@nmfmt{\NAT@nm}%
   \NAT@spacechar\NAT@@open\if*#1*\else#1\NAT@spacechar\fi\NAT@hyper@{\NAT@date}}
  {}{}

\makeatother

\title{BDH-CQ: In-Context Learning with Recurrent Latent Reasoning}

\author[1]{Björn Engdahl}
\author[1]{Adrian Kosowski}
\author[1]{Jan Chorowski}
\author[1]{Zuzanna Stamirowska}
\author[1]{Przemysław Uznański}
\author[1]{Junlin Jiang}
\author[1]{Rohan Phadke}
\author[2]{Remigiusz Kinas}
\author[3]{Richard Zhong}

\affil[1]{Pathway, \url{https://pathway.com/research}}
\affil[2]{Bielik AI \url{https://bielik.ai/}}
\affil[3]{New York University}

\begin{document}

\maketitle

\begin{abstract}
We introduce \model, a reasoning model that combines in-context learning with recurrent latent
reasoning. Inputs presented at inference time continuously update the model's recurrent memory;
the model then solves a query through iterative computation in a high-dimensional latent space,
without verbalizing its intermediate reasoning.  We evaluate the model on the public
\arcone{} evaluation set and use controlled ARC-like interventions to study what it learns from
demonstrations, how consistently it applies an inferred transformation, and which concepts remain
difficult.  A 150M-parameter configuration reaches 29.5\% pass@2 at a computed inference cost of
\$0.0007 per task---less than one-tenth of a cent.  This operating point breaks through the
previously reported ARC\nobreakdash-AGI\nobreakdash-1 cost--accuracy Pareto frontier, establishing a new state of the art
in benchmark cost efficiency.
\end{abstract}

\section{Introduction}

In-context learning allows a model to acquire a new skill from examples presented at inference
time \citep{brown2020language}.  In autoregressive language models, chain-of-thought (CoT) prompting complements this
capability with a computational workspace: demonstrations specify what to do, while generated
intermediate tokens support the computation required to do it
\citep{nye2021workscratchpadsintermediatecomputation,wei2022chain}.  Reinforcement learning on verifiable problems has made
this combination increasingly powerful, eliciting long reasoning traces, self-verification, and
adaptive solution strategies \citep{deepseek2025r1}.  It also couples reasoning to serial
narration.  As reasoning traces have grown, so have token consumption, latency, and inference
compute.  Every intermediate state must be projected through a discrete vocabulary, emitted
autoregressively, and consumed again before computation can continue.

Latent reasoning opens a different computational regime.  Instead of verbalizing every
intermediate result, a model repeatedly transforms its continuous hidden state and decodes only the
answer.  This removes natural-language tokens as the mandatory representation of internal
computation and allows a state to preserve partial hypotheses or candidate transformations without
serializing each one.  Continuous-thought and recurrent-depth models show that this form of computation allows for advanced reasoning by exploring multiple solution paths in parallel
\citep{hao2024coconut,geiping2025recurrent,saunshi2025looped}.  The separation is also consistent
with evidence that human conceptual reasoning and language rely partly on distinct systems
\citep{fedorenko2016language}.

Yet latent reasoning and in-context learning have largely developed separately. Chain-of-thought reasoning
language models learn flexibly from context but usually allocate additional computation through
generated tokens.  Compact recursive solvers reason iteratively in latent space, but their ARC
pipelines learn evaluation tasks through optimization and task-specific identities
\citep{wang2025hrm,jolicoeur2025trm,arcprize2025hrmanalysis}. They do not acquire an unseen
transformation solely from demonstrations during inference.

We introduce \model, a reasoning system that brings these capabilities together.  Demonstrations
of a previously unseen task update recurrent memory; the query is then solved through iterative
computation in a high-dimensional latent workspace.  Intermediate reasoning states
are not decoded into language.  \model{} makes memory, adaptation, and inference part of the same
computational fabric. Neither task identifiers nor evaluation-task demonstration pairs participate in training, and no parameters
are updated at inference time. A 150M-parameter configuration reaches 29.5\% pass@2 on ARC-AGI-1
at a computed \$0.00070 per task, breaking through the previously reported cost--accuracy Pareto
frontier.

We use the Abstraction and Reasoning Corpus (ARC) as both an evaluation and an experimental
substrate \citep{chollet2019measure}.  ARC specifies new visual transformations through sparse
demonstrations and requires exact outputs, making learning and rule consistency directly
inspectable.  We use the term \emph{demonstration-conditioned operator schema} for the behavior
that results when demonstrations bind a reusable visual operation for the current task.  The
controlled experiments test what such schemas can bind, how far they extrapolate, which ones compose with one another, and where their execution fails.

Our contributions are:
\begin{itemize}
  \item We introduce \model, which combines in-context learning through evolving recurrent memory
  with iterative reasoning in a structured continuous latent space.
  \item We establish a new state-of-the-art point on the public \arcone{} cost--accuracy frontier:
  29.5\% pass@2 at a computed \$0.00070 per task, and describe the ARC-style data mixture
  used to train the system.
  \item We map capabilities and failure modes across concept-organized transformation families,
  separating isolated correct outputs from consistent rule application.
  \item We use controlled ARC-like tasks to test what concepts the new reasoning system can learn
  and express from demonstrations.
\end{itemize}

We close by outlining how the same architecture can be extended through model scaling, longer
training, language and mathematical reasoning, constraint satisfaction, and harder ARC tasks.

The paper first describes the system and its training and then presents the evaluation and
behavioral analyses. The behavioral analysis of the evaluation sets is presented in the appendix.

\section{ARC as a controlled, verifiable testbed for BDH-CQ’s in-context generalization}

ARC was designed to study \emph{skill-acquisition efficiency}: not only whether a system possesses
a skill, but how much experience and prior structure it requires to acquire one
\citep{chollet2019measure}.  Each task specifies a new transformation through only a handful of
demonstrations.  The system must infer which objects, relations, and operations matter to derive a transformation rule and apply the resulting rule to a new input.  ARC therefore operationalizes the fast generalization that
in-context learning is intended to provide.

This task format also makes the behavior of \model{} accessible to controlled analysis.  The
demonstrations constitute the task specification, while the small colored grids require little
factual knowledge or linguistic fluency.  They can nevertheless express object relations,
counting, symmetry, topology, spatial transformations, and compositions.  Every answer is exactly
verifiable, every failure is visually inspectable, and multiple test inputs reveal whether a rule
is applied consistently rather than guessed once \citep{chollet2024arcprize}.

\begin{figure}[t]
  \centering
  \includegraphics[trim={0 1cm 0 0}, clip, width=\linewidth]{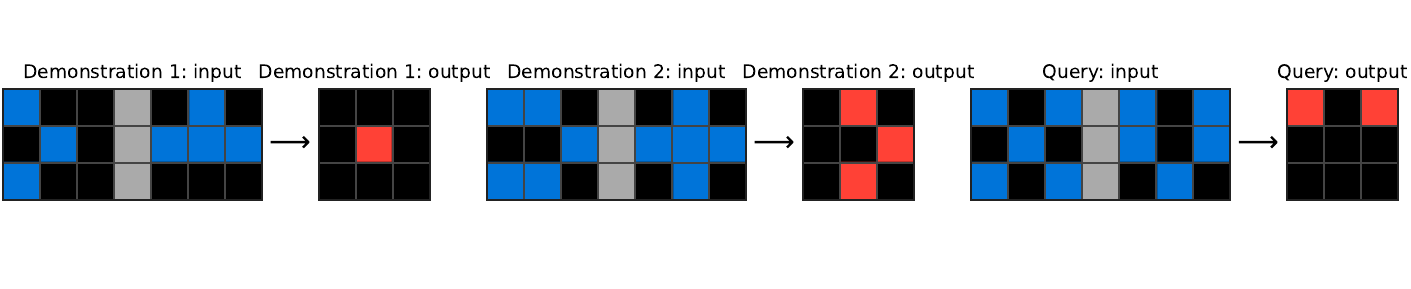}
  \caption{An ARC-AGI-1 training task (identifier 0520fde7).  Two of three demonstrations and the
  held-out query are shown; the query output is included here for explanation.  The task is compact,
  exact, and easy to inspect, yet its transformation must be inferred from examples.}
  \label{fig:arc-example}
\end{figure}

Figure~\ref{fig:arc-example} illustrates this task.  The input contains two binary panels
separated by a gray column.  From the demonstrations, the system must infer that the output marks
the cells occupied in both panels and transfer that relation to the query.  Nothing in the input
names intersection, alignment, or the meaning of the output color.

ARC gives us a controlled visual language in which learning from context, latent computation,
exact execution, and inference cost can be studied together.

\section{Introducing BDH-CQ: In-context learning through recurrent memory and latent reasoning}

\subsection{BDH provenance}

We have previously introduced the Dragon Hatchling (BDH), a post-Transformer sequence-model
architecture built around high-dimensional positive activations, low-rank communication, and a
recurrent associative state \citep{kosowski2025bdh}.  Its GPU-oriented formulation uses BDH layers
combining ReLU-low-rank transformations with linear attention in a large neuron or feature space.
The design is brain-inspired but not brain-imitative.  It draws on principles biology got right,
namely local interaction, sparse activity, persistent state, and continual adjustment, and applies
them to a modern sequence model.  That work established BDH's language, sequence-modeling, and
interpretability properties.

BDH layers also support recurrent systems for constraint satisfaction.  Our Sudoku system
iteratively refines a visual state until it satisfies the puzzle constraints
\citep{pathway2025sudoku}.  \model{} extends
this line into a new reasoning system that combines a structured latent workspace and recurrent
computation over model depth with an interface for learning visual transformations from
demonstrations.  We use ``BDH'' for the architectural family and ``\model'' for the system
introduced here.  The complete evaluated system includes input transformations, candidate
construction, ranking, and the inference pipeline.

\subsection{In-context learning through recurrent memory}

Let a task provide demonstrations $D=\{(x_t,y_t)\}_{t=1}^{K}$ and query input $x^\star$.  Rather
than compressing the demonstrations into a single task vector, \model{} processes their elements
sequentially.  At a high level, its recurrent memory evolves as
\begin{equation}
  S_t = U_\theta(S_{t-1}, D_t),
  \label{eq:memory}
\end{equation}
where $D_t$ denotes the content of the $t$-th demonstration and $\theta$ remains fixed.  Thus,
information available to later inputs depends on associations accumulated from earlier inputs and
outputs.  This recurrent contextual state plays a role analogous to the context-dependent
associations constructed by attention, while avoiding a growing explicit key--value cache.
The interpretation is generally related to attention, fast-weight memory, and linear-attention views of
contextual association
\citep{vaswani2017attention,ba2016fastweights,katharopoulos2020linear,geva2021keyvalue}, with linear attention being the conceptually simplest standalone realization of linear correction rules on $S$, capturing the special case $S_t = S_{t-1} + U_\theta(D_t)$.

\subsection{Recurrent latent reasoning}

After the demonstrations and query have been ingested, \model{} performs iterative computation in
a structured latent workspace $H_r$. This happens after the ingestion of the $K$ demonstrations into memory. We summarize this process through the following equations which capture the encoding of inputs $x^\star$, latent reasoning, and decoding of the output $\hat y$.

\begin{align}
  H_0 &= E_\theta(x^\star, S_K), \\
  H_{r+1} &= F_\theta(H_r, S_K), \quad r=0,\ldots,R-1, \\
  \hat y &= G_\theta(H_R).
  \label{eq:reasoning}
\end{align}

The contextual memory $S_t$ and reasoning workspace $H_r$ have different conceptual roles.
$S_t$ changes as evidence is encountered and supports in-context learning.  $H_r$ carries the
ongoing computation used to answer the current query.  This organization defines the system-level
interface studied in this paper.  Dimensions, exact update rules, and implementation details
remain proprietary.

\section{Training data and objective}

\subsection{ARC task formulation}

An ARC task $T_i$ contains $K_i$ demonstration pairs with $Q_i$ test pairs, $Q_i\ge 1$.
\begin{equation}
  T_i = \left(\{(x_{i,j},y_{i,j})\}_{j=1}^{K_i},
  \{(x^{\mathrm{test}}_{i,q},y^{\mathrm{test}}_{i,q})\}_{q=1}^{Q_i}\right).
\end{equation}
During training, the model predicts outputs after preceding examples have been incorporated into
recurrent context.  At a high level, the objective trains the system to use preceding examples and
produce exact target grids.  We report the task interface and data provenance; the complete
internal training recipe remains proprietary.

\subsection{Training mixture}

First, we train a 150M-parameter model on a curated collection of ARC-style data.  The dataset
combines privately curated examples with publicly available data from the ARC-AGI-1 training
set~\citep{chollet2019measure}, RE-ARC~\citep{hodel2024addressing},
ConceptARC~\citep{moskvichev2023conceptarc}, ARC-Heavy~\citep{li2025combining}, and
ARC-GEN100K~\citep{moffitt2025arcgen}.  We apply additional augmentations to increase the variety
of the training data.

\section{Public ARC-AGI-1 evaluation}

We evaluate on the 400-task public \arcone{} evaluation split
\citep{chollet2019measure,chollet2024arcprize}. The default system configuration follows the
ARC-AGI leaderboard's two-attempt convention (pass@2), producing up to two ranked candidates, receiving a score if either of the candidates is correct.  The reported dollar value is
computed from measured hardware time for the leaderboard's score--cost plane.  Other plotted
systems use the costs reported by the leaderboard, which may represent hardware estimates or API
prices.  We additionally report pass@1 and test-pair accuracy when analyzing behavior.

\begin{figure}[t]
  \centering
  \includegraphics[width=\linewidth]{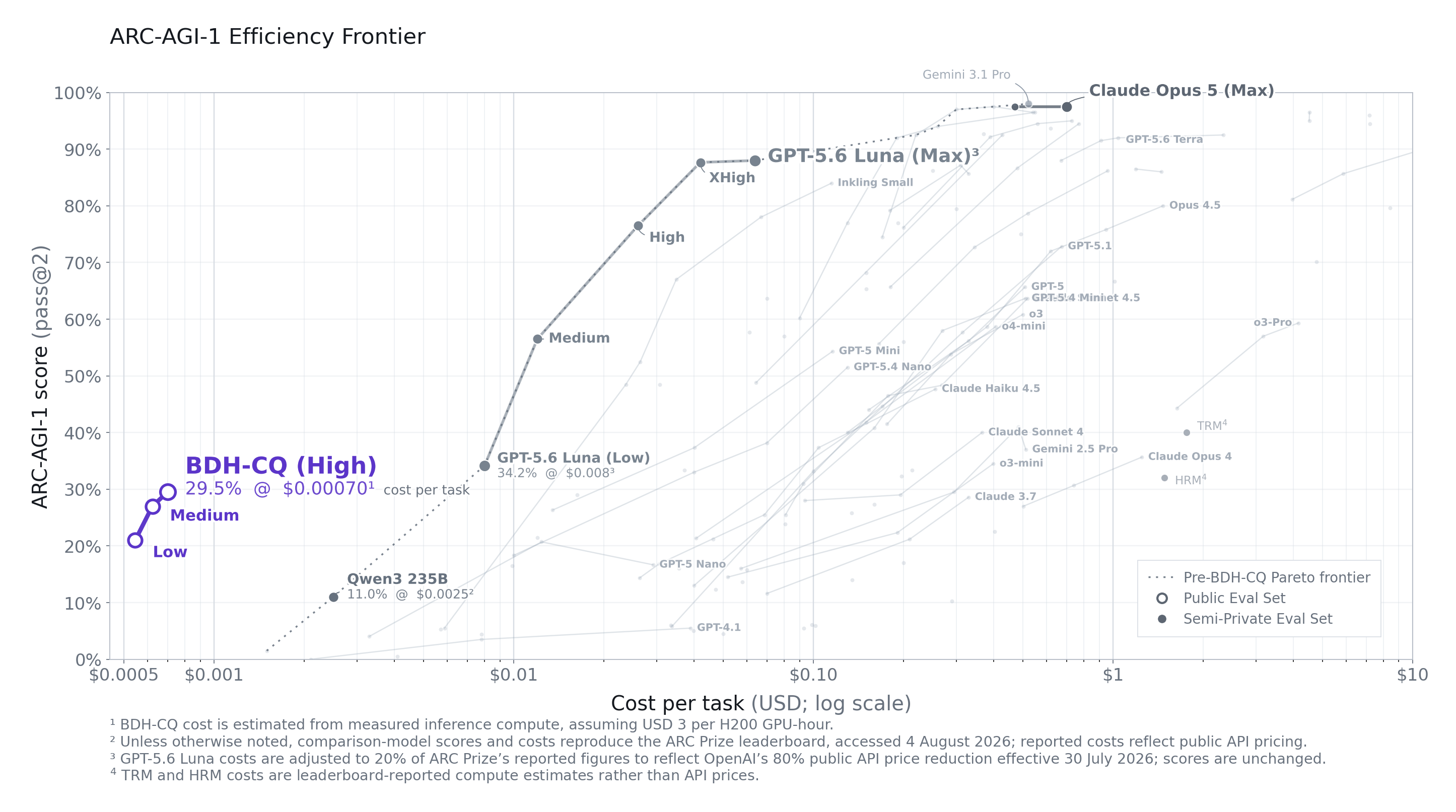}
  \caption{ARC-AGI-1 score versus computed cost per task. Data points were collected from the official ARC Prize leaderboard on August 4, 2026. The mark at the left margin indicates
  the 29.5 pass@2. The point breaks through the previously reported cost--accuracy Pareto
  frontier and establishes a new state of the art in ARC-AGI-1 cost efficiency.
  }
  \label{fig:cost-frontier}
\end{figure}

At the default 
point, the 150M-parameter system reaches 29.5\% pass@2 in approximately
0.85 H200 GPU-seconds per task.  At \$3 per H200-hour this gives a computed cost of
\$0.00070 per task, less than one-tenth of a cent.  As Figure~\ref{fig:cost-frontier} shows, this
point lies beyond the previously reported Pareto frontier: no plotted system attains at least this
accuracy at equal or lower reported cost.  This establishes a new state of the art in ARC-AGI-1
cost efficiency.  Using ARC Prize's reported costs as of July 2026, BDH-CQ is approximately
57x cheaper than GPT 5.6 Luna (Low) which scores 34.2\% at \$0.040 \citep{arcprize2026leaderboard}. When accounting for OpenAI’s 80\% public API price reduction of GPT 5.6 Luna on July 30, 2026, which is not reflected in ARC Prize’s data as of August 6, 2026, BDH-CQ is approximately 11x cheaper than GPT 5.6 Luna (Low).

\paragraph{Independent evaluation.}

An independent black-box audit conducted by co-authors from Bielik and New York University
reproduced the deployed system's 29.5\% pass@2 score on the public \arcone{} evaluation set.
The auditors evaluated the system under a documented
protocol without access to model weights.
Their report also evaluates BDH-CQ on ConceptARC and a hand-crafted evaluation set
\citep{independent2026audit, independent2026audit2}.

\section{Behavioral analysis on ARC-like tasks}
We analyze \model's performance across two different sets: ConceptARC \citep{moskvichev2023conceptarc} in section \ref{arc-concept} and the official public evaluation set with curated tasks matching the distribution of the public evaluation set (see Appendix).

\begin{table}[t]
  \caption{Headline results on the public ARC-AGI-1 evaluation and on ConceptARC.
  Intervals are
  descriptive Wilson 95\% intervals.}
  \label{tab:behavioral-headline}
  \centering
  \small
  \begin{tabular}{@{}llrrr@{}}
    \toprule
    Set / condition & Unit & $N$ & pass@1 & pass@2 [95\% Wilson] \\
    \midrule
    ARC-AGI-1 public & tasks & 400 & 97 (24.25\%) & 118 (29.50\%) [25.24, 34.15] \\
                        & test pairs & 419 & 108 (25.78\%) & 130 (31.03\%) \\
    ConceptARC, semantic IDs & tasks & 160 & 73 (45.63\%) & 95 (59.38\%) [51.63, 66.68] \\
                        & test pairs & 480 & 332 (69.17\%) & 374 (77.92\%) \\
    ConceptARC, opaque IDs & tasks & 160 & 72 (45.00\%) & 96 (60.00\%) [52.26, 67.27] \\
                        & test pairs & 480 & 334 (69.58\%) & 374 (77.92\%) \\
    \bottomrule
  \end{tabular}
\end{table}

\subsection{Concept-organized capability profile}
\label{arc-concept}

ConceptARC organizes tasks into 16 families, providing a structured basis for
examining which visual transformations the deployed system can learn.  Each family
contains ten tasks and thirty test inputs.  We use strict task accuracy as the
primary measure: a task is counted as correct only when all three test inputs are
solved.  Test-pair accuracy is reported alongside it to distinguish consistent
application of a transformation from success on individual inputs. We treat each concept family as a probe of whether demonstrations can bind the corresponding visual operator schema.

Performance varies substantially across concept families (Table~\ref{tab:concept-profile}).
In the opaque-identifier replication, \texttt{ExtendToBoundary},
\texttt{FilledNotFilled}, and \texttt{TopBottom2D} each reach 9/10 
pass@2, whereas \texttt{Copy} and \texttt{Order} reach 2/10.  With only ten
tasks per family, however, these counts are a profile rather than a reliable
ranking: the descriptive Wilson intervals for 9/10 and 2/10 overlap broadly
($59.6\%$--$98.2\%$ and $5.7\%$--$51.0\%$, respectively).

Several families show a substantial difference between test-pair and strict-task
accuracy.  \texttt{Copy}, for example, reaches 19/30 semantic test-pair
pass@2 but only 2/10 strict tasks (20/30 pairs in the opaque replication).  The
system can therefore produce correct outputs for many inputs in this family, but
does not yet apply the inferred transformation consistently across all inputs of
a task.

\begin{figure}[t]
  \centering
  \includegraphics[width=\linewidth]{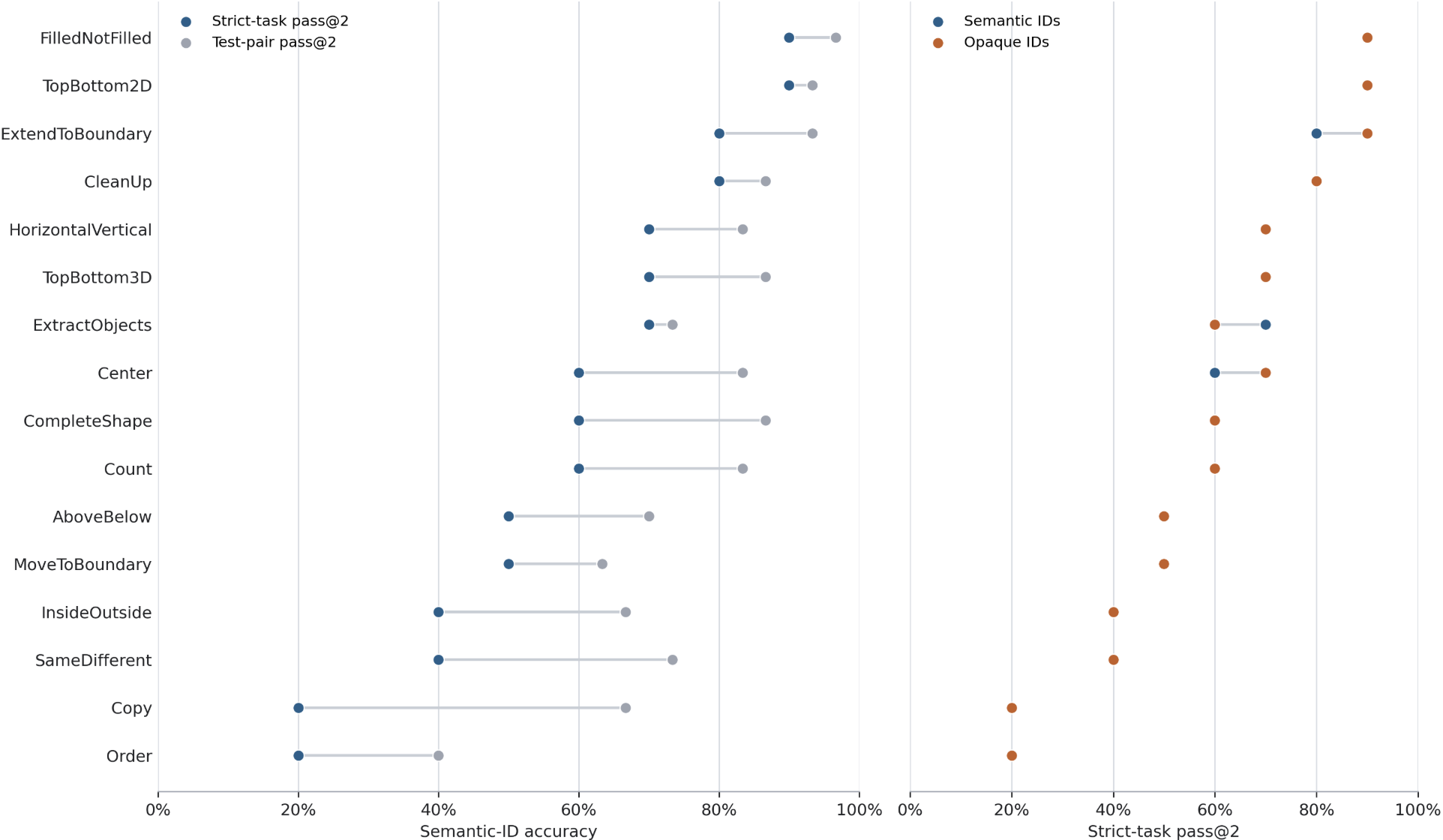}
  \caption{ConceptARC pass@2 by concept area.  Left: semantic-ID
  test-pair and strict-task accuracy; their gap captures correct outputs that do
  not transfer across all three test inputs of a task.  Right: strict-task
  accuracy when the same grids are sent with semantic identifiers in
  concept-grouped batches or opaque identifiers in concept-mixed batches.  The
  profile is nearly unchanged; this combined intervention does not isolate the
  effects of identifiers and batch composition separately.}
  \label{fig:behavioral-conceptarc-profile}
\end{figure}

\begin{table}[t]
  \caption{ConceptARC profile by concept area.  Each area has ten tasks and
  thirty test pairs.  Pass@2 is shown for both the semantic and
  opaque-identifier executions; pass@1 and test-pair columns are from the
  semantic execution.  The three task-pass@2 changes are shown in bold.}
  \label{tab:concept-profile}
  \centering
  \small
  \begin{tabular}{@{}lrrrrr@{}}
    \toprule
    Concept area & \shortstack{Task p@2\\semantic} & \shortstack{Task p@2\\opaque}
      & \shortstack{Task\\p@1} & \shortstack{Pair\\p@1} & \shortstack{Pair\\p@2} \\
    \midrule
    \texttt{FilledNotFilled}     & 9 & 9 & 8 & 27 & 29 \\
    \texttt{TopBottom2D}         & 9 & 9 & 7 & 26 & 28 \\
    \texttt{ExtendToBoundary}    & 8 & \textbf{9} & 7 & 27 & 28 \\
    \texttt{CleanUp}             & 8 & 8 & 7 & 23 & 26 \\
    \texttt{HorizontalVertical}  & 7 & 7 & 4 & 21 & 25 \\
    \texttt{TopBottom3D}         & 7 & 7 & 7 & 25 & 26 \\
    \texttt{ExtractObjects}      & 7 & \textbf{6} & 5 & 19 & 22 \\
    \texttt{Center}              & 6 & \textbf{7} & 3 & 18 & 25 \\
    \texttt{CompleteShape}       & 6 & 6 & 4 & 21 & 26 \\
    \texttt{Count}               & 6 & 6 & 2 & 20 & 25 \\
    \texttt{AboveBelow}          & 5 & 5 & 4 & 20 & 21 \\
    \texttt{MoveToBoundary}      & 5 & 5 & 5 & 19 & 19 \\
    \texttt{InsideOutside}       & 4 & 4 & 2 & 16 & 20 \\
    \texttt{SameDifferent}       & 4 & 4 & 4 & 20 & 22 \\
    \texttt{Copy}                & 2 & 2 & 2 & 19 & 20 \\
    \texttt{Order}               & 2 & 2 & 2 & 11 & 12 \\
    \midrule
    Total                         & 95 & 96 & 73 & 332 & 374 \\
    \bottomrule
  \end{tabular}
\end{table}

\subsection{Simple operators extrapolate; ordering and nesting expose distinct boundaries}

The ConceptARC profile shows where performance differs across broad concept
families, but not how an individual operation behaves as its demands increase.
We therefore construct controlled tasks after freezing the model and vary one
source of complexity at a time.  The experiments carry forward boundary
propagation as a relative strength, copying and ordering as pressure points, and
nested containment as a direct test of increasing relational depth.

We generated fresh ARC-like tasks after freezing the model.  Every task contains
three demonstrations and two held-out inputs, and every target output is produced
by a deterministic oracle.  Four families vary one factor while holding the rule
fixed: horizontal propagation distance, the number of independent motif copies,
the number of bars that must be sorted, and the number of nested containment
relations.  Demonstrations use propagation distances 1--3, one or two copies,
sequences of length 2--4, and nesting depths 1--3.  Held-out instances extend
beyond those ranges.  We use six tasks per level for propagation and copying and,
after an initial pilot, 18 per level around the ordering and nesting transitions
(12 for the two longest ordering levels).

\begin{figure}[ht]
  \centering
  \includegraphics[width=0.75\linewidth]{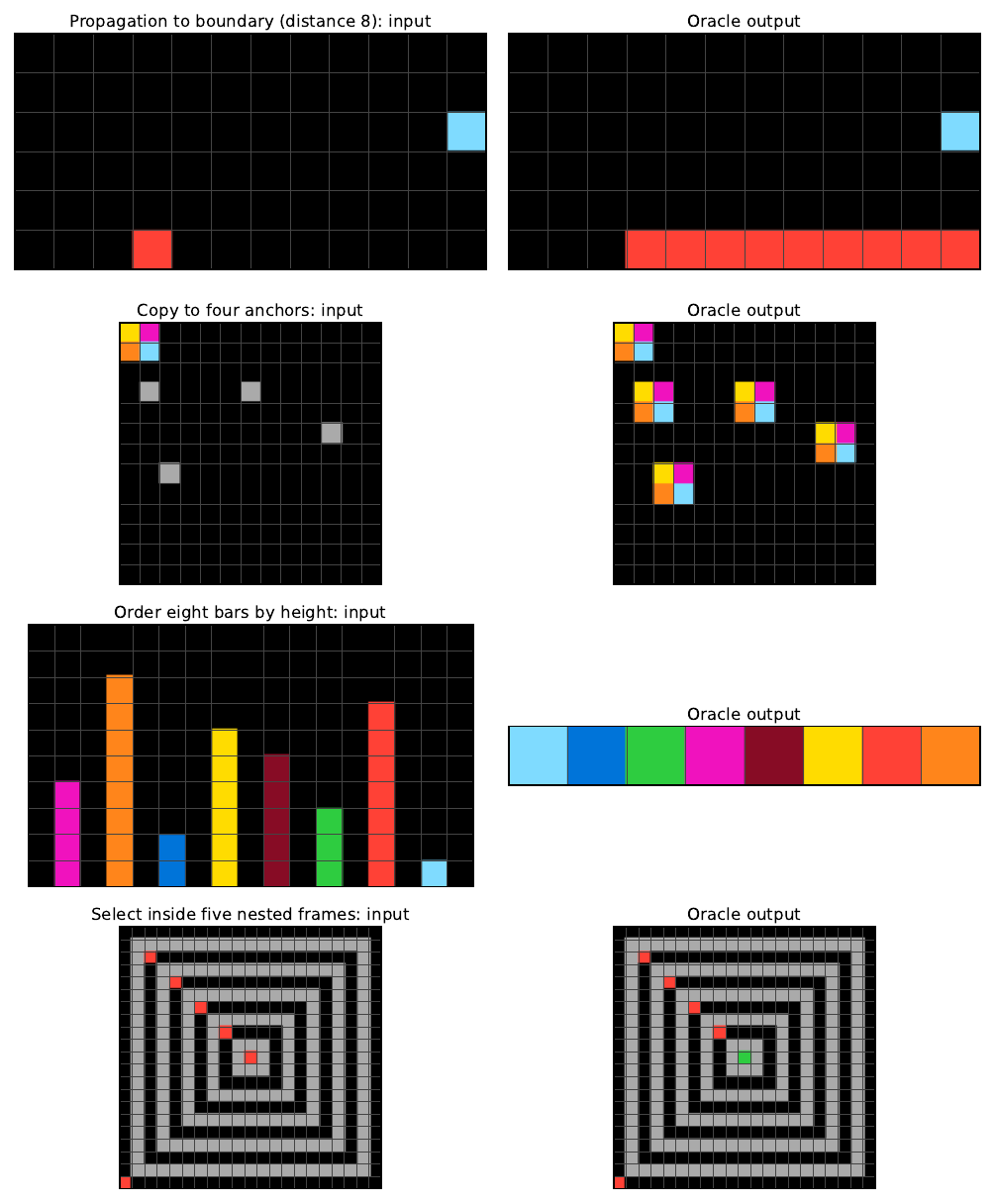}
  \caption{Representative held-out examples from the four controlled
  generalization families.  From top to bottom: extend a seed to the boundary,
  copy a motif to every gray anchor, order bar colors from shortest to tallest,
  and recolor only cells inside every nested frame.  These examples show the
  transformations whose distance, multiplicity, sequence length, and nesting
  depth are varied in Figure~\ref{fig:controlled-scaling}.}
  \label{fig:scaling-examples}
\end{figure}

Figure~\ref{fig:scaling-examples} shows one held-out input and oracle output from
each family at the upper end of the tested range.  Propagation and copying enlarge
a locally specified operation; ordering requires constructing a sequence from
eight spatially separated objects; and nesting requires resolving five containment
relations before recoloring the selected cells.

\begin{figure}[ht]
  \centering
  \includegraphics[width=\linewidth]{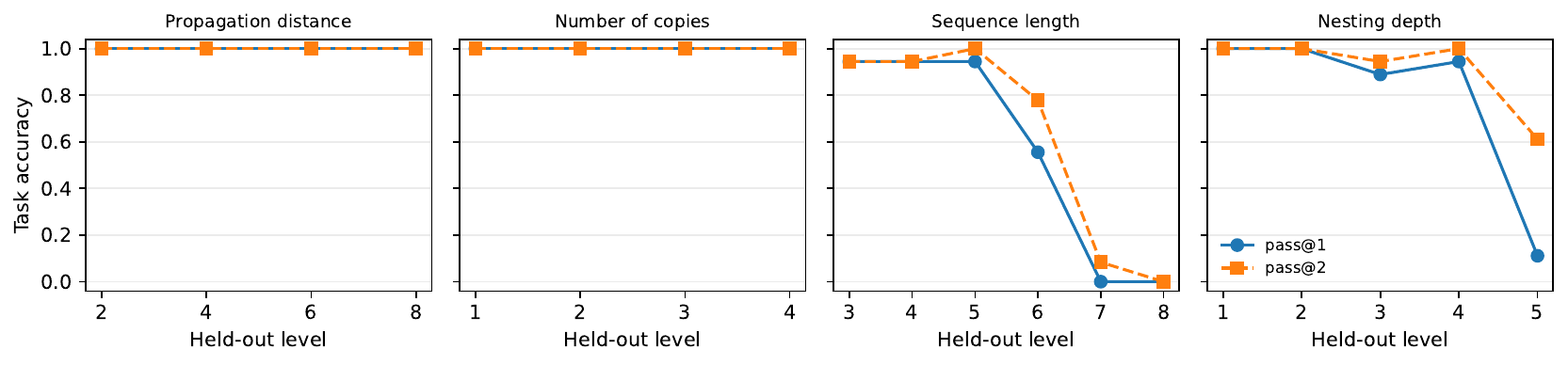}
  \caption{Controlled generalization curves reporting exact held-out-output
  accuracy.  Ordering and nesting include a larger preregistered replication
  around the transition; propagation and copying use 12 outputs per point,
  ordering and nesting use 36, and the two longest ordering levels use 24.}
  \label{fig:controlled-scaling}
\end{figure}

Figure~\ref{fig:controlled-scaling} shows sharply different generalization
patterns.  Propagation remains correct on 48/48 held-out outputs at both pass@1
and pass@2 across distances 2--8, and copying remains correct on 48/48 as the
number of target sites increases from one to four; neither reaches a ceiling in
the tested range.  Ordering is nearly saturated through five objects, then falls
to 29/36 outputs at length six, 8/24 at seven, and 1/24 at eight at pass@2.
Nesting remains nearly saturated through depth four but falls to 29/36 at depth
five.

The two declines have different signatures.  At ordering length eight, only 3/24
held-out outputs have the correct dimensions.  By contrast, all 36 depth-five
nesting outputs have the correct dimensions, and mean best-candidate cell
accuracy exceeds 99.9\%; errors typically differ from the target in a single
containment decision.  Ordering failures therefore affect construction of the
output as a whole, whereas nesting failures usually preserve its structure and
make a localized relational error.

To determine if the source of the model's failure is its execution capacity or its extrapolation ability, we rerun byte-identical
length-eight ordering and depth-five nesting inputs under  contexts of varying complexity.  A
``short'' context stops at length or depth three to four; a ``supported'' context
includes one demonstration at the test complexity.  As
Table~\ref{tab:matched-support} shows, matched support raises depth-five nesting
from 19/24 to 24/24 exact outputs at pass@2.  For ordering, it recovers 13/24
outputs from a 0/24 baseline.  The nesting cliff is therefore largely a failure
to extrapolate the demonstrated relation depth.  Long ordering also benefits
from support, but retains an execution bottleneck.

\begin{table}[ht]
  \caption{Output accuracy depends on coverage in the in-context examples.  The
  same 24 held-out inputs are evaluated under two contexts while every test pair
  remains byte-identical.  ``Short'' demonstrations stop below the test
  complexity; ``supported'' adds one demonstration at the target length or
  depth.  p@1 uses the first candidate and p@2 accepts either of two candidates;
  entries report exact outputs out of 24.}
  \label{tab:matched-support}
  \centering
  \small
  \begin{tabular}{llcc}
    \toprule
    Family & Context & Output p@1 & Output p@2 \\
    \midrule
    Ordering (length 8) & Short & 0/24 & 0/24 \\
    Ordering (length 8) & Supported & 12/24 & 13/24 \\
    Nesting (depth 5) & Short & 15/24 & 19/24 \\
    Nesting (depth 5) & Supported & 16/24 & 24/24 \\
    \bottomrule
  \end{tabular}
\end{table}

\subsection{Context binds dense mappings, but composition is operation-dependent}

To probe contextual memory directly, each task defines a fresh color permutation through its demonstrations and
requires elementwise application to two held-out sequences.   The system solves all 96 held-out outputs at rank one as the number
of simultaneous bindings increases from two to eight (24/24 at every level).
The model therefore recovers and consistently applies a relatively dense
task-specific mapping introduced entirely through context.

\begin{figure}[ht]
  \centering
  \includegraphics[width=0.52\linewidth]{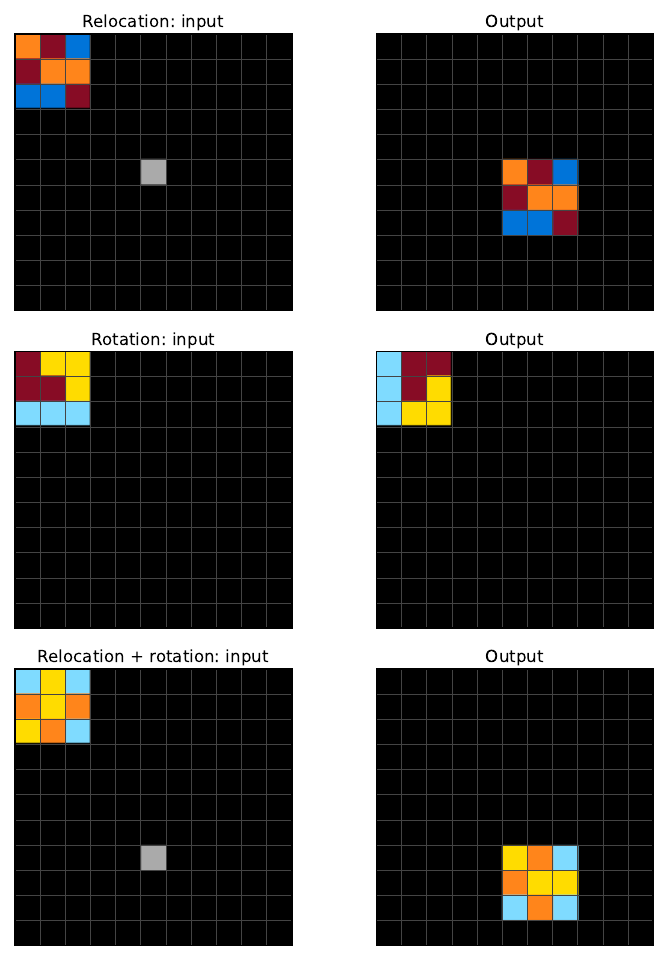}
  \caption{Representative held-out pairs from an independent motif family.
  Relocation moves the motif to the gray anchor; rotation changes it in place;
  their composition performs both.}
  \label{fig:composition-examples}
\end{figure}

To determine the ability of the model to compose operations, we compare the model's performance on a new experiment where the demonstration examples are either a single operation or a composition of operations. Each demonstration and test input board consists of a 3x3 grid of colors, which we call a motif, placed in the top left corner, as shown in Figure \ref{fig:composition-examples}. We test horizontal reflection, a demonstration-
defined swap of two colors, and clockwise rotation alone. Additionally, we also test demonstration examples which consist of the composition of each one of these operations with relocation of the motif to a marked position.  Three
dense, asymmetric 3$\times$3 motif families provide 72 held-out outputs per
condition.  One family, the original family, has a fixed color layout, where different motifs in the family differ only by a permutation of colors. There are also two shuffled families, where motifs do not follow a fixed color layout, in order to balance color multiplicities.  Across families, we hold the
canvas, upper-left source position, single-cell anchor, demonstrations per task,
and semantic operations constant.

Relocation alone, atomic reflection, and atomic rotation are each solved on
72/72 held-out outputs.  Rotation composed with relocation is solved on 72/72 tasks, while reflection composed with relocation is solved on 47/72 tasks.  The model learned to compose reflection with relocation on most color layouts, but not all.  Color swap is acquired atomically only in the original family (26/72
pooled) and the model never learns to compose it with relocation (0/72).  The two shuffled families each reach only
1/24 in isolation, showing that the fixed layout in the original motif family made the swap
substantially easier to infer; their composed failures cannot be attributed to
composition alone.  Figure~\ref{fig:composition-examples} illustrates the task
construction, and Table~\ref{tab:composition-controls} reports the pooled results.

\begin{table}[ht]
  \caption{Exact held-out-output pass@2 pooled across three dense 3$\times$3
  motif families (72 outputs per condition).  The canvas, source and anchor
  locations, and semantic operations are held fixed across families.  Relocation
  is the standalone baseline; subsequent rows compare each operation alone with
  its composition with relocation.}
  \label{tab:composition-controls}
  \centering
  \begin{tabular}{lcc}
    \toprule
    Operation & Alone & Composed with relocation \\
    \midrule
    Relocation & 72/72 & n/a \\
    \midrule
    Reflection & 72/72 & 47/72 \\
    Rotation & 72/72 & 72/72 \\
    Color swap & 26/72 & 0/72 \\
    \bottomrule
  \end{tabular}
\end{table}

\subsection{Within-task consistency}

The 18.5-point gap between semantic ConceptARC pair accuracy (77.92\%) and
strict task accuracy (59.38\%) is not just a consequence of a stricter metric.
Under pass@2, the system solved zero test pairs for 13 tasks, one for
15 tasks, two for 37 tasks, and all three for 95 tasks.  Thus, 52/160 tasks have
one or two correct test inputs but are not solved as tasks.  Under a
rule-induction account of ARC, a correctly induced rule should transfer to every
test input of its task; this partial success is evidence that the observed
transformation is not applied consistently across inputs.

\subsection{Identifier and batch-context replication}

The most direct request-side confound is that the system could exploit a semantic
\texttt{task\_id} or a concept-grouped batch context rather than the grids.  The
opaque replication replaced identifiers with cryptographically opaque labels and
mixed concept areas within batches.  Aggregate performance did not move: it
solved 374/480 test pairs in both conditions and 96/160 opaque tasks versus
95/160 semantic tasks.  On the paired test-pair outcome, there were exactly six
semantic-only and six opaque-only pass@2 successes.  This supplies no directional
evidence that removing those request-side cues changed performance.

Aggregate stability did not imply identical outputs.  Across the two ConceptARC
executions, 442/480 first candidates, 276/480 complete ordered candidate lists,
and 455/480 attempt counts agreed.  Request context, call time, or an internal
search path can therefore change the candidates returned without changing the
aggregate score.  The replication rules out this combined identifier-and-batch
confound as an explanation for the observed ConceptARC score; it does not make
ConceptARC a fresh benchmark, rule out exposure through training or checkpoint
selection, or isolate the two interventions factorially.

\subsection{Attempt delivery, repeatability, and effort}

In the opaque ConceptARC replication, all 75 single-candidate
records were already correct at rank one.  Consequently, nominal pass@2 should
not be interpreted as the result of two independently sampled attempts for every
input.

On the full public ARC-AGI-1 evaluation, the \textsc{MIN} effort setting cost
one third of \textsc{standard} ($\$0.00088399$ versus $\$0.00265246$ per task)
and scored 111/400 rather than 118/400  pass@2, a difference of $-1.75$
percentage points.  The paired split was 105 tasks solved by both settings, 13
standard-only, 6 min-only, and 276 neither (two-sided exact McNemar $p=0.167$).
All four reported endpoints favor \textsc{standard} in point estimate, but this
comparison is statistically unresolved.

Repeated identical requests were byte-identical at both effort tiers: the
standard re-run matched all 419 ARC-AGI-1 test inputs from the earlier standard
run, and repeated min runs also matched all 419.

\section{Scaling reasoning effort increases pass@2}
In this section we evaluate the impact of reasoning effort (latent thinking effort) on the pass@2 and cost.

\begin{figure}
  \centering
  \includegraphics[width=300pt]{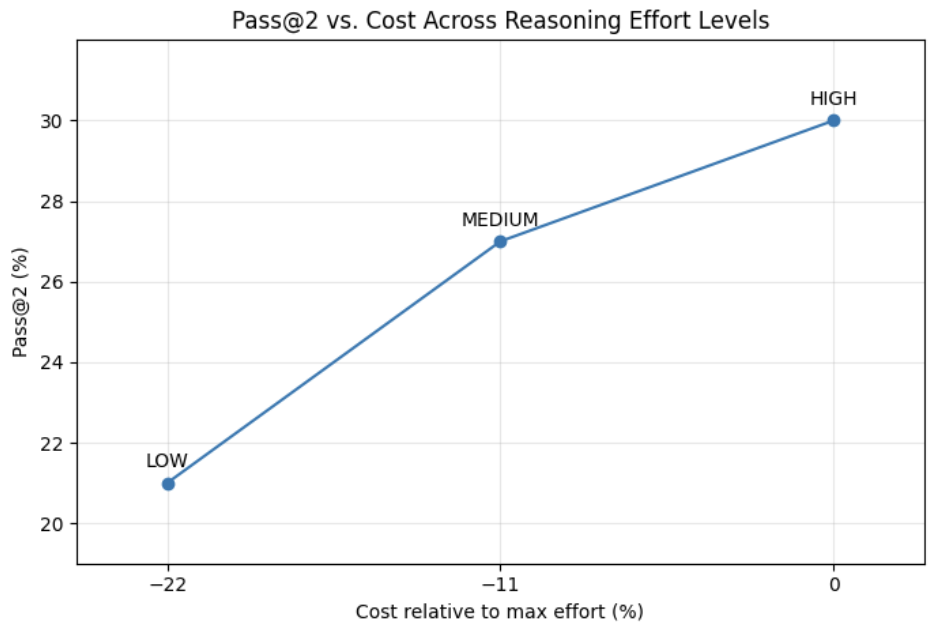}
  \caption{Showing how pass@2 and cost scale with reasoning effort.}
  \label{fig:effort}
\end{figure}

For this task, we train the model changing the levels of latent reasoning during training. In doing this, we get a model
that has been exposed to different reasoning efforts during training. For inference, we can choose the level of reasoning
to apply. In table \ref{tab:effort} and figure \ref{fig:effort}, we show an increase in pass@2 for different levels of latent reasoning (three different reasoning effort levels). It is
evident that adding more effort from LOW to MEDIUM and from MEDIUM to HIGH boosts the accuracy. 

\begin{table}[t]
  \caption{Comparing pass@2 and cost across reasoning efforts LOW, MEDIUM, HIGH}
  \label{tab:effort}
  \centering
  \small
  \begin{tabular}{llcccc}
    \toprule
    Effort & Pass@2 & Cost reduction \\
    \midrule
    HIGH & 29.5\% & 0\% \\
    MEDIUM & 27\% & 11\% \\
    LOW & 21\% & 22\%  \\
    \bottomrule
  \end{tabular}
\end{table}

\section{Related work}

\paragraph{Verbalized reasoning.}
Chain-of-thought prompting elicits intermediate natural-language steps and can substantially improve
reasoning in sufficiently large language models \citep{wei2022chain}.  This mechanism combines
in-context specification with an autoregressive computational scratchpad.  \model{} combines the
same contextual flexibility with a continuous recurrent workspace, making intermediate language
generation unnecessary.

\paragraph{Continuous thoughts in language models.}
Coconut feeds a Transformer's previous final hidden state back as the next input embedding and uses
a curriculum that progressively replaces verbal CoT steps with continuous thoughts
\citep{hao2024coconut}.  It provided the seminal demonstration of continuous hidden-state feedback
in an autoregressive language model and showed that one state can preserve multiple candidate
continuations during planning.  Its staged internalization curriculum is essential to the reported
performance; curriculum controls in the same study recover much of the result on its logical
reasoning tasks.  \model{}
explores a different operating regime: demonstrations update recurrent task memory, and a
structured multi-vector workspace solves previously unseen visual transformations.  We evaluate
that combination as a complete costed system and probe its binding, extrapolation, composition, and
execution behavior.

Theory gives a more precise motivation than a generic appeal to higher-dimensional states.
For directed graph reachability, \citet{zhu2025superposition} show that continuous thoughts encode multiple search frontiers simultaneously and can expand them in parallel.  Their construction
demonstrates the central computational opportunity: one continuous state can carry several active
hypotheses.  \citet{xu2026formal} characterize a broader trade-off: latent iteration can exploit
parallel structure in computation graphs, whereas stochastic token decoding supplies natural
mechanisms for approximate counting and sampling.

\paragraph{Compressed discrete and communicated latent states.}
Abstract-CoT replaces verbal rationales with a short autoregressive sequence from a learned reserved
vocabulary and reports up to 11.6-fold fewer reasoning tokens while approaching verbal-CoT
post-training performance \citep{ramji2026abstract}.  It occupies the discrete end of the
nonlinguistic-reasoning spectrum: compressed and abstract, yet still serial and externally decoded.
Separately, latent communication transfers embeddings, hidden states, or inference caches between
agents rather than between reasoning steps within one model \citep{liu2026latentcommunication}.
This adjacent literature reinforces the need to distinguish continuous latent thought, compact
abstract-token languages, and inter-agent latent channels.

\paragraph{Recurrent depth and looped models.}
Recurrent-depth language models repeatedly apply a shared block to a sequence-wide latent state
and can improve some reasoning benchmarks with additional iterations
\citep{geiping2025recurrent}.  Looped Transformers provide empirical and theoretical evidence that
many iterative problems require effective depth without requiring distinct parameters at every
depth, and can simulate bounded tokenized CoT computations under stated constructions
\citep{saunshi2025looped,zhu2026scalinglatentreasoninglooped, zhang2025softthinkingunlockingreasoning}. These works primarily treat recurrent-depth as a way to design parameter-efficient architectures and to scale test-time compute. Our focus is demonstration-dependent learning and expression of visual
transformations.

\paragraph{Task-trained recursive solvers.}
HRM and TRM recursively update latent and candidate-answer states and report strong results on
Sudoku, mazes, and ARC \citep{wang2025hrm,jolicoeur2025trm}.  Their ARC pipeline is transductive:
demonstration pairs from evaluation tasks are augmented and used in optimization. Each augmented
puzzle receives a learned identity embedding, and predictions are voted over augmentations
\citep{arcprize2025hrmanalysis}.  A previously unseen hidden task therefore requires backward-pass
adaptation before it can be evaluated. BDH-CQ targets that specific conjunction: task information
is written into recurrent memory through context, and latent computation applies it without
puzzle-specific optimization. ARC Prize reports costs of \$1.48 per task for HRM and
\$1.76 per task for TRM, reflecting their coupling of task-specific optimization with inference
\citep{arcprize2025hrmanalysis,arcprize2026leaderboard}. HRM-Text subsequently demonstrated that
hierarchical recurrence can also support efficient conditional language modeling at one billion
parameters, with strong results on knowledge, reading, and mathematical benchmarks
\citep{wang2026hrmtext}.  

\paragraph{ARC and concept-organized evaluation.}
ARC was proposed to study skill-acquisition efficiency under sparse evidence
\citep{chollet2019measure}.  ConceptARC groups ARC-like tasks around a designed ontology of spatial
and semantic concepts, enabling analyses hidden by one aggregate score
\citep{moskvichev2023conceptarc}.  We use this ontology as the organizing map for capability
analysis.
\section{Discussion}

The system-level result is direct: a 150M-parameter \model{} reaches 29.5\% pass@2 on the public
\arcone{} evaluation set at a computed \$0.00070 per task.  This breaks through the reported
cost--accuracy Pareto frontier and establishes a new state of the art in benchmark cost efficiency.
The unusually low starting cost also creates room to increase model capacity and reasoning compute
while remaining competitive with existing systems.

The controlled experiments show what that combination can express.  \model{} applies dense
task-specific color mappings to all 96 held-out outputs, demonstrating substantial contextual
binding capacity.  Boundary propagation and copying extrapolate throughout the tested ranges,
while ordering and nested containment expose distinct limits.  Matched demonstrations eliminate
the observed depth-five nesting failures and substantially improve length-eight ordering, showing
that coverage in the in-context examples can determine whether a learned operation extrapolates.
Composition is more representation-dependent: rotation composes with relocation on all 72
held-out outputs, whereas reflection succeeds on 47/72 and varies sharply across motif families;
color swapping is not acquired reliably outside the original fixed layout.  The resulting picture
is not a single scalar notion of reasoning difficulty, but a capability profile shaped by
contextual binding, demonstrated support, output construction, and the representation through
which operations are composed.

The present study establishes this account for ARC-like visual reasoning.  The ConceptARC ontology
organizes the learned capability profile, while post-freeze generators test generalization and
localize current boundaries in consistency, composition, conditional execution, and output
construction.

\subsection{Outlook}

\model{} points toward a broader family of BDH-based reasoning systems in which memory,
adaptation, and inference are part of the same computational fabric.  The immediate direction is
to scale \model{} itself.  The current result comes from a compact model and leaves a large cost
budget for increasing capacity while retaining a favorable operating point.  Larger models and
longer training runs will test whether BDH's favorable sequence-model scaling carries into
in-context latent reasoning and whether the capability boundaries identified here move
predictably with scale.

The second direction is breadth.  ARC-AGI-2 is the next visual-reasoning target
\citep{chollet2025arcagi2}; the present diagnostics provide a concrete development agenda around
output construction, conditional binding, demonstration coverage, and multi-operator composition.
Sudoku and other constraint-satisfaction domains offer complementary tests of long-horizon latent
refinement \citep{pathway2025sudoku}.  Language and mathematical reasoning will test whether the
same recurrent memory can acquire tasks from textual demonstrations while retaining BDH's
sequence-model capabilities.  Because BDH layers support language modeling as well as latent
reasoning, future systems may also combine continuous internal computation with verbalized steps
when communication, verification, or tool use requires them.

Finally, \model{} keeps the strengths of language models, but pushes beyond token-by-token processing
toward parallel latent reasoning.  BDH-CQ is one instance of this direction rather than its
endpoint.  Because BDH layers support language modeling as well as latent visual reasoning, future
BDH-based systems could combine continuous and verbalized chains of thought in one recurrent
architecture. Combining such systems could retain the bandwidth and efficiency of latent computation while
decoding intermediate language when communication, verification, or tool use benefits from it.
This direction is motivated by evidence that language and nonlinguistic reasoning are partly
separable in humans \citep{fedorenko2016language}, without assuming that the models reproduce the
corresponding biological mechanisms.

\subsection{Scaling model size}

The \model{} architecture scales naturally to large model sizes, admitting tensor sharding patterns inherited  from the BDH architecture that make it particularly easy to train at 1T scale. Early experiments confirm Transformer-like scaling laws apply during pretraining at scales from 1B to 600B parameters, while preserving the latent reasoning capabilities specific to \model{}.

\section{Conclusion}

\model{} demonstrates that in-context learning and recurrent latent reasoning can coexist in a
compact, practical system.  Demonstrations modify recurrent memory at inference time, and the
resulting task is solved through iterative continuous computation rather than a verbalized chain of
thought.  On ARC-AGI-1, this design reaches 29.5\% pass@2 for \$0.00070 per task, establishing a new
state of the art in benchmark cost efficiency.  The controlled experiments further show that the
system binds new mappings, extrapolates several learned operators, benefits predictably from
demonstration coverage, and exhibits structured rather than uniform limits under composition.
Together, these results establish in-context latent reasoning as a concrete path toward systems
that learn new tasks from context while reasoning beyond the token stream.

\bibliography{references}
\newpage
\appendix

\section{Evaluation set analysis}

\subsection{Evaluation sets and surface descriptors}

The concept-organized experiments above isolate individual transformations and structural
demands.  We complement them by profiling the public \arcone{} evaluation set, a generated set
calibrated to match its measured distribution, and a generated set stratified by required
operation. We report this quantity as exact whole-task solve rate rather than test-pair accuracy.

Table~\ref{tab:evalset-overview} summarizes the three cohorts.  

\begin{table}[t]
  \caption{Exact whole-task solve rate under STANDARD effort for the evaluation cohorts used in the
  profiling analysis. Solve rate refers to pass@2.  The generated cohorts serve different experimental purposes and are not
  directly comparable.}
  \label{tab:evalset-overview}
  \centering
  \small
  \begin{tabular}{lrrr}
    \toprule
    Evaluation cohort & Tasks & Solved & Solve rate \\
    \midrule
    Public \arcone{} evaluation & 400 & 118 & 29.5\% \\
    Calibrated generated & 400 & 149 & 37.2\% \\
    Mechanic-stratified generated & 1,131 & 337 & 29.8\% \\
    \bottomrule
  \end{tabular}
\end{table}

Simple surface properties provide little explanation for success.  We evaluate 39 bucketings of
grid size, color count, object count, and demonstration count on the public set and 38 on the
calibrated generated set.  Nine public-set bucketings exceed the permutation search null, while
none does so on the generated set.  Grid size is the strongest individual descriptor but remains
weak, with reported pseudo-$R^2=0.072$ on public tasks and $0.010$ on generated tasks.  Thus, the
measured differences in success cannot be reduced to one coarse surface descriptor.

\subsection{Transformation mechanics}

We next group tasks by the operation required to solve them, which we call the task's
\emph{mechanic}.  Unconstrained generation, without controlling the requested transformation mechanic, produces roughly 80 rotation tasks for every three
gravity-and-stacking tasks, leaving rare mechanics too sparsely sampled to estimate.  We therefore
generate approximately balanced samples across 16 mechanics, yielding 1,131 tasks.  The resulting
profile is shown in Table~\ref{tab:mechanic-profile}.

\begin{table}[t]
  \caption{Exact whole-task solve rate on the mechanic-stratified generated set.  Wilson
  confidence intervals and mean grid side are reported for each generated mechanic.}
  \label{tab:mechanic-profile}
  \centering
  \small
  \begin{tabular}{lrrrr}
    \toprule
    Mechanic & $n$ & Solve rate & 95\% CI & Mean side \\
    \midrule
    Flood fill & 51 & 68.6\% & $[55, 80]$ & 15.6 \\
    Denoising & 65 & 56.9\% & $[45, 68]$ & 18.8 \\
    Scaling & 82 & 53.7\% & $[43, 64]$ & 15.1 \\
    Cropping and extraction & 74 & 44.6\% & $[34, 56]$ & 19.0 \\
    Translation & 87 & 35.6\% & $[26, 46]$ & 16.6 \\
    Tiling and repetition & 79 & 34.2\% & $[25, 45]$ & 19.5 \\
    Line drawing & 84 & 29.8\% & $[21, 40]$ & 16.6 \\
    Rotation & 75 & 25.3\% & $[17, 36]$ & 14.9 \\
    Recolor by property & 75 & 25.3\% & $[17, 36]$ & 17.0 \\
    Object sorting and rank & 68 & 19.1\% & $[12, 30]$ & 14.9 \\
    Object counting & 86 & 18.6\% & $[12, 28]$ & 16.7 \\
    Reflection & 72 & 16.7\% & $[10, 27]$ & 17.7 \\
    Symmetry completion & 61 & 16.4\% & $[9, 28]$ & 25.5 \\
    Occlusion repair & 56 & 16.1\% & $[9, 28]$ & 24.3 \\
    Panel set operation & 48 & 10.4\% & $[5, 22]$ & 11.0 \\
    Gravity and stacking & 68 & 2.9\% & $[1, 10]$ & 14.6 \\
    \bottomrule
  \end{tabular}
\end{table}

Mechanic is a useful diagnostic stratifier for observed solve rate in this generated set. The solve rates span 65.7 percentage points, from 68.6\% for flood fill to 2.9\% for gravity and
stacking.  In 20,000 permutations of solved labels at fixed group sizes, the 95th-percentile
spread is 27.0 points and no permutation matches the observed spread ($p<0.00005$).  At the
category level, mean grid side has essentially no linear
association with mechanic solve rate ($r=-0.022$), and panel set operations have the smallest mean
side but the second-lowest solve rate.  This aggregate correlation does not control for size
variation within mechanics.

This profile is not an intrinsic ranking of the operations.  Every stratified task was authored
by GPT-5.6 from a prompt naming the target mechanic, so each rate also reflects how that generator
instantiates the requested operation.  Public and generated mechanic rates do not show a reliable
rank correspondence ($\rho=0.300$ across 16 mechanics).  Hand-written comparison families expose
a substantial task-construction confound: hand-written gravity tasks score 80--100\%, compared
with 2.9\% for generated gravity-and-stacking tasks, and hand-written counting tasks score
85--95\%, compared with 18.6\% for generated object counting.  Touching-panel overlays remain
difficult under both authoring modes, scoring 7.5\% when hand-written and 10.4\% when generated.
The mechanic table therefore characterizes this evaluation distribution, while controlled
generators are required to isolate properties of the operation itself.

\subsection{Controlled structural manipulations}

The mechanic taxonomy describes \emph{which} transformation a task requires, but does not encode
how demonstrations select or parameterize that transformation.  We construct ten deterministic
Python-generated ladders to vary such properties directly.  Each step holds the grid at
$12\times12$, uses three demonstrations, one test pair, and a fixed palette, and contains 40
tasks.  The retained analysis contains 1,840 task evaluations.  The selection results use a
corrected 240-task rerun because the original generator omitted a required rule from some
demonstration sets.  Table~\ref{tab:evalset-ladders} summarizes the largest observed changes.

\begin{table}[t]
  \caption{Selected controlled-ladder results.  Each condition contains 40 tasks, except the
  pooled two-rule and unseen-value conditions, which contain 120.  Each comparison is specific to
  one hand-written puzzle family and has not yet been tested across operations.  Changes are in
  percentage points.}
  \label{tab:evalset-ladders}
  \centering
  \small
  \begin{tabular}{lrrr}
    \toprule
    Manipulation & Reference & Changed condition & Change \\
    \midrule
    Conditional rule selection & 100.0\% & 56.7\% & $-43.3$ \\
    Parameter absent from demonstrations & 30.0\% & 0.0\% & $-30.0$ \\
    Panel union, two to three panels & 65.0\% & 2.5\% & $-62.5$ \\
    Support chain, shortest to eight objects & 80.0\% & 27.5\% & $-52.5$ \\
    \bottomrule
  \end{tabular}
\end{table}

For conditional selection, a corner marker determines which of two demonstrated rules applies.
A control in which the marker varies but both values invoke the same rule is solved on 40/40
tasks.  Pooling color, position, and count cues, genuine selection between two rules is solved on
68/120 tasks, or 56.7\% ($95\%$ CI $[48, 65]$), a decrease of 43.3 points
($p=9\times10^{-9}$, two-sided Fisher's exact test).  The control shows that the decrease attaches
to using the cue to select a rule, rather than to the presence of a varying marker.

A second ladder uses marker count to specify a downward shift.  When the test value appears among
the demonstrations, the model solves 12/40 tasks.  When the value is absent, it solves 0/120:
0/40 for an unseen value interpolated between demonstrated values and 0/80 for two extrapolation
conditions ($p=1.4\times10^{-8}$, two-sided Fisher's exact test, pooled against the
demonstrated-value condition).  Within this family, the relevant boundary is whether the value is
demonstrated, not whether it lies inside the demonstrated range.  The demonstrated-value condition
itself reaches only 30.0\%, so the result may share the
conditional-parameterization limitation observed in rule selection.

For panel union, two panels at opposite corners are solved on 26/40 tasks, whereas three panels are
solved on 1/40.  The generator holds answer color and expected union density fixed across panel
counts.  Separation matters even before a third panel is added: two touching panels score 3/40,
while opposite-corner panels score 26/40.  This pattern is consistent with a segmentation
limitation, although the analysis does not directly test the model's segmentation.

Other manipulations show no detectable cost over the tested range.  Four axis-aligned operations
in sequence are solved on 38/40 tasks, counting 10--12 objects reaches 34/40 compared with 36/40
for one to three objects, and three independently moving objects are solved on 40/40.  Within the
axis-aligned family, operation count carries no detectable cost over the tested range.  Dependency
between objects behaves differently: support chains decrease monotonically from 80.0\% with one
object resting on another
to 67.5\%, 52.5\%, and 27.5\% as the chain grows to three, five, and eight objects.  The
three-object independent control shows that multiplicity alone does not cause failure at that
scale.  Object count and dependency depth co-vary in the longer chains, so the gradient is
consistent with a dependency cost but does not isolate it from object count at every level.

\subsection{Failure structure and scope}

Finally, we compare incorrect predictions on the 400-task public and calibrated generated sets.
The calibrated set contributes 251 failed tasks and the public set 282.  As
Table~\ref{tab:evalset-failures} shows, most failures preserve output dimensions and palette, and
shape-correct failures often differ from the target in a small fraction of cells.

\begin{table}[t]
  \caption{Properties of incorrect predictions on the calibrated generated and public sets.
  Median cell error is computed only among failures with the correct output dimensions.
  Insufficient target-edit coverage is an output-only descriptor: the prediction modifies fewer
  target-change cells than required; it does not identify an inferred rule.}
  \label{tab:evalset-failures}
  \centering
  \small
  \begin{tabular}{lrr}
    \toprule
    Failure property & Calibrated generated & Public \\
    \midrule
    Output dimensions correct & 89.2\% & 89.7\% \\
    Palette correct & 72.9\% & 78.0\% \\
    Median cell error, shape correct & 4.0\% & 8.3\% \\
    Input reproduced verbatim & 8.4\% & 4.6\% \\
    Insufficient target-edit coverage & 15.5\% & 7.8\% \\
    \bottomrule
  \end{tabular}
\end{table}

These output-level signatures characterize the final grids but do not identify the latent rule
the model followed.  The evaluator returns grids rather than a reasoning trace, so a correct rule
applied incompletely is observationally indistinguishable from a narrower rule applied completely.

The generated analyses have additional scope limitations.  Manual review found at least one task
whose stated test output contradicts the rule expressed consistently by its demonstrations; the
prevalence of this defect is unknown.  Independent LLM mechanic labels agree with the requested
label on 82.9\% of the nonrandom 659-task subset that was labeled, and some mechanic definitions
overlap.  Deduplication removed 24\% of the merged generated pool, concentrated in flood fill and
panel set operation; additional batches produced few new distinct forms at these extremes.  Each
ladder condition contains only 40 tasks from one puzzle family.  Consequently, the results
establish strong within-family differences for conditional selection, unseen parameter values,
and panel number and separation, together with a graded association for support-chain length; they
do not yet establish that these differences generalize across visual operations.

Taken together, the evaluation set analyses separate surface appearance from transformation and
structural demand.  Coarse descriptors such as grid size weakly predict success; operation-based
groups reveal large differences but remain confounded by task authoring; and deterministic
ladders isolate sharper limitations in selecting and parameterizing demonstrated rules.  Many
incorrect outputs preserve output dimensions, and among shape-correct failures many differences
are localized.  These outputs are consistent with substantial partial recovery, although final grids
alone cannot establish where the underlying reasoning failed.

\end{document}